\documentclass[11pt, letterpaper]{article}

\usepackage{amsmath}
\usepackage{amssymb}
\usepackage{amsthm}
\usepackage{graphicx}
\usepackage{hyperref}
\usepackage{url}
\usepackage{algorithm}
\usepackage{algorithmic}
\usepackage{caption}
\usepackage{subcaption}
\usepackage{booktabs}
\usepackage{authblk}
\usepackage{appendix}
\usepackage[utf8]{inputenc}
\usepackage[T1]{fontenc}
\usepackage[english]{babel}
\usepackage{geometry}
\usepackage{xcolor}
\usepackage{fancyhdr}
\usepackage{microtype}
\usepackage{parskip}
\usepackage{tabularx}
\usepackage{enumitem}

\hypersetup{
    colorlinks=true,
    linkcolor=blue,
    filecolor=magenta,      
    urlcolor=cyan,
    citecolor=red,
}

\definecolor{headercolor}{RGB}{0, 50, 100}
\title{Enhancing LLMs for Impression Generation in Radiology Reports through a Multi-Agent System}

\author[1]{Fang Zeng}
\author[1]{Zhiliang Lyu}
\author[1]{Quanzheng Li}
\author[1]{Xiang Li}

\affil[1]{Department of Radiology, Massachusetts General Hospital and Harvard Medical School}

\begin{document}

\maketitle

\begin{abstract}
This study introduces "RadCouncil," a multi-agent Large Language Model (LLM) framework designed to enhance the generation of impressions in radiology reports from the finding section. RadCouncil comprises three specialized agents: 1) a "Retrieval" Agent that identifies and retrieves similar reports from a vector database, 2) a "Radiologist" Agent that generates impressions based on the finding section of the given report plus the exemplar reports retrieved by the Retrieval Agent, and 3) a "Reviewer" Agent that evaluates the generated impressions and provides feedback. The performance of RadCouncil was evaluated using both quantitative metrics (BLEU, ROUGE, BERTScore) and qualitative criteria assessed by GPT-4, using chest X-ray as a case study. Experiment results show improvements in RadCouncil over the single-agent approach across multiple dimensions, including diagnostic accuracy, stylistic concordance, and clarity. This study highlights the potential of utilizing multiple interacting LLM agents, each with a dedicated task, to enhance performance in specialized medical tasks and the development of more robust and adaptable healthcare AI solutions. 
\end{abstract}

\section{Introduction}
In radiology workflow, radiologists traditionally interpret imaging studies and manually draft detailed reports, including an "impression" section that summarizes clinically significant findings and possible diagnosis, which is a vital part of the report for referring physicians and patient care. This process is time-consuming and subject to variability impacted by the radiologist’s knowledge and experience \cite{mussurakis1996observer}. Automated impression generation has the potential to improve report consistency, reduce radiologist workload, and enhance the overall quality of radiology reports ~\cite{sun2023evaluating}. Such a feature is especially needed with the recent growth in the demands for medical imaging, which are straining radiologists, leading to possible burnout and impacting their ability to provide timely and precise reports \cite{maskell2022does}.

Large Language Models (LLMs) have shown exceptional capabilities in understanding and generating text that is coherent and contextually relevant, making them promising tools for auto-generating impressions from findings in radiology reports. A few studies have investigated LLMs’ ability for the impression summarization task \cite{sun2023evaluating,builtjes2024evaluating,zhang2024constructing,ma2024iterative,tie2024personalized,liu2023tailoring}, demonstrating the potential of LLMs to revolutionize radiology workflow by automating the report generation process. Various techniques, such as prompt engineering, model fine-tuning, and retrieval-augmented generation (RAG) \cite{ma2024iterative}, have been used to improve their performance and mitigate the limitations of LLMs, including hallucinations and insufficient domain knowledge. However, current approaches often lack the ability to integrate these techniques together within a unified framework and the scalability to incorporate new techniques and/or requirements in impression generation. Furthermore, applying LLM, even with domain adaptation techniques, for text generation without reflecting and reviewing the output of the model risks introducing hallucinations that could compromise the reliability of the results and lead to serious errors in clinical decision-making \cite{omiye2024large,haltaufderheide2024ethics}.   

Recently, multi-agent systems have emerged as a promising approach for various natural language processing tasks \cite{wang2024survey}. By leveraging the strengths of multiple specialized agents, each focusing on specific tasks, a multi-agent system can enhance the overall quality and reliability of the model performance. In addition, techniques that are used to enhance the performance of a single LLM can be readily applied to each agent and the whole system. Thanks to the advanced natural language and code/structured data processing capability of LLMs, multi-agent systems can understand model inputs in generalizable forms, write and execute codes, and perform reasoning to orchestrate the collaborative process, effectively "wrapping" any models in a more intelligent and robust form. With efficient memory management among agents, the multi-agent system can mitigate the limitations of the context window size, thus improving the handling of extensive and complex tasks for more accurate and relevant outputs. Furthermore, through the collaboration among LLM agents, the multi-agent system can combine diverse capabilities from models with different domain expertise, potentially leading to emergent properties \cite{de2023emergent}. Some preliminary attempts have been made to test multi-agent frameworks in healthcare settings, such as clinical trial outcome prediction \cite{yue2024ct}, diagnosis from case reports \cite{ke2024enhancing}, medical education \cite{wei2024medco}, and medical question answering \cite{tang2023medagents,smit2023we}. 

In this study, we develop a multi-agent framework named “RadCouncil” to generate the impression section in radiology reports based on the finding section, using reports from chest X-ray imaging as the study case. RadCouncil consists of three specialized agents: (1) a Retrieval agent that searches for similar reports from an external database, (2) a Radiologist agent that generates the impression section from the finding section with reference to the retrieved reports, and (3) a Review agent that evaluates the generated impression and provides feedback to the Radiologist agent with possible revisions. The main contribution of this work is twofold: RadCouncil presents an innovative multi-agent approach that mimics the workflow in radiology clinical practice, incorporating key steps of report writing, reference to documents, and review by self/peers. Secondly, the demonstrated effectiveness of RadCouncil in analyzing and generating specialized medical text shows the potential of multi-agent systems across a broad spectrum of medical applications, extending beyond radiology to other healthcare domains where complex task coordination and interplay among different roles are crucial. 

\section{Methodology}
\begin{figure}[h]
\centering
\includegraphics[width=0.9\textwidth]{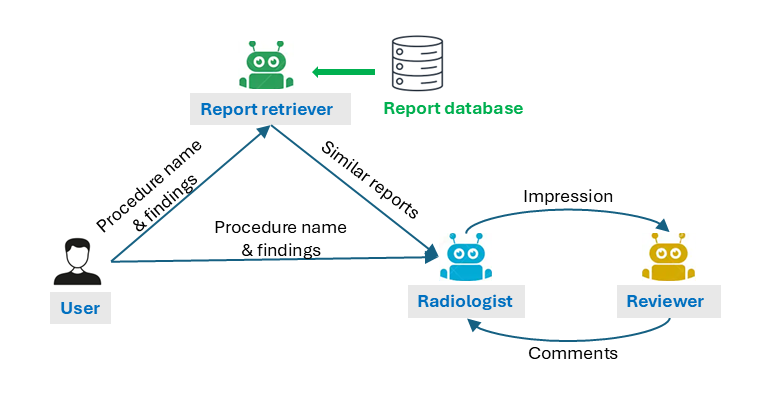}
\caption{Diagram of the proposed RadCouncil framework, illustrating the interactions between the three agents, the report database, and the user. }\label{fig1}
\end{figure}

\subsection{RadCouncil: Overview and Workflow}
In this work, we introduce a multi-agent system for generating radiology report impressions, leveraging both the generative power of LLMs and established medical knowledge from existing reports. The system employs three specialized agents: a Report Retriever that uses vector similarity matching to find relevant historical reports, a Radiologist agent that generates impressions with reference to retrieved examples, and a Reviewer agent that ensures consistency between findings and generated impressions. This architecture not only helps maintain accuracy but also reduces hallucinations by grounding the generation process in real medical examples and cross-checking by multiple LLMs. An illustrative example of the RadCouncil's workflow is shown in Table 1.
\begin{table}[h!]
\centering
\caption{Sample workflow of the RadCouncil system}
\label{tab:multi_agent_system}
\scriptsize 
\begin{tabular}{|p{2.5cm}|p{12cm}|}
\hline
\textbf{User} & \textbf{PROCEDURE NAME: XR CHEST PORTABLE. FINDINGS:} \newline
Lines/tubes: ET tube tip is projected at the level of the carina with tip projected near the origin of the right mainstem bronchus. Evidence of embolization coils project over the right upper hemithorax. 
Lungs: Patchy ill-defined opacities in the right upper lung are again noted. No gross pulmonary edema. 
Pleura: There is no pleural effusion or pneumothorax. 
Heart and mediastinum: The heart and the mediastinum are unchanged. 
Bones: The thoracic skeleton is unchanged. \\ \hline

\textbf{Retriever} & \textbf{Here are some examples:} \newline
\textbf{EXAMPLE 1:} \newline
PROCEDURE NAME: XR CHEST PORTABLE. FINDINGS: 
Lines/tubes: Endotracheal tube terminates below the carina in the proximal right mainstem bronchus. 
Lungs: Redemonstration of patchy ill-defined opacities in the right mid to upper lung. Focus of patchy opacity in the left base is unchanged. No pulmonary edema. 
Pleura: There is no pleural effusion or pneumothorax. 
Heart and mediastinum: The heart and the mediastinum are unchanged.
Bones: The thoracic skeleton is unchanged. \newline
\textbf{IMPRESSION:} \newline
Endotracheal tube terminates in the proximal right mainstem bronchus.
Unchanged patchy opacities in the right mid to upper lung representing pulmonary hemorrhage. Superimposing aspiration or pneumonia may also be considered. \newline
\textbf{EXAMPLE 2:} \newline
$\dots$ \\ \hline

\textbf{Radiologist} & \textbf{IMPRESSION:} \newline
Tubes as described above. \newline
Patchy ill-defined opacities in the right upper lung likely reflect atelectasis, although concurrent aspiration and/or pneumonia cannot be excluded by radiograph. \newline
Unchanged mild pulmonary edema not seen on this study, but previously noted. \newline
No pleural effusion or pneumothorax. \\ \hline

\textbf{Reviewer} & \textbf{Upon reviewing the 'FINDINGS' and 'IMPRESSION' sections, I noticed an inconsistency:} \newline
* In the 'FINDINGS' section, it is stated that "No gross pulmonary edema" is seen, whereas in the 'IMPRESSION' section, it is mentioned that there is "unchanged mild pulmonary edema not seen on this study, but previously noted." This implies that pulmonary edema was present in a previous study, but it is not consistent with the current findings. \newline
\textbf{REVISION: YES} \\ \hline

\textbf{Radiologist} & \textbf{IMPRESSION:} \newline
Tubes as described above. \newline
Patchy ill-defined opacities in the right upper lung likely reflect atelectasis, although concurrent aspiration and/or pneumonia cannot be excluded by radiograph. \newline
No evidence of pulmonary edema or pleural effusion/pneumothorax. \\ \hline

\textbf{Reviewer} & \textbf{Upon reviewing the 'FINDINGS' and 'IMPRESSION' sections, I did not notice any inconsistencies:} \newline
* The description of tubes in the 'IMPRESSION' section is consistent with the 'FINDINGS' section. \newline
* The patchy ill-defined opacities in the right upper lung are described consistently in both sections. \newline
* The absence of pulmonary edema and pleural effusion/pneumothorax is also described consistently in both sections. \newline
\textbf{REVISION: NO}\\ \hline
\end{tabular}
\end{table}
\subsubsection{Report Retriever}
The Report Retriever agent retrieves similar reports from an external database implemented by a vector database. Using the procedure name and findings provided by the input, the report retriever finds the matching exemplar reports by converting the input into its vector representation and searching the external radiology report database based on similarity score (e.g., cosine similarity). It will then fetch the top \textit{k} similar vectors, where \textit{k} is a predefined parameter balancing the retrieval efficiency and comprehensiveness.
\subsubsection{Radiologist}
The Radiologist agent is responsible for generating the impression based on the procedure name and findings provided by the user. This agent utilizes the exemplar reports retrieved by the Report Retriever to augment its initial generation, improving the written style, the significance of the findings, and their interpretation. If a revision request is made by the Reviewer agent, the Radiologist agent will revise the previously generated impression based on the reviewer's comments and feedback. 
\subsubsection{Reviewer}
The reviewer agent is designed to examine the consistency between the generated impression and the provided findings, ensuring that the generated impression aligns with each of the findings. If any inconsistencies are identified, the reviewer requests revisions from the radiologist, leading to the next iteration of the workflow. Otherwise, if no inconsistencies are found, the reviewer confirms that no revisions are needed, and the workflow ends with the generated impression as the final output. While other types of errors, such as missing clinically significant findings or misinterpretations, may exist, our study specifically addresses the potential consistency errors. Checking and revising the consistency error could also significantly reduce the hallucinations in the generated impressions, enhancing the trustworthiness of the system. 
\subsection{Radiology Report Database}
The radiology report database serves as an external resource to support impression generation with exemplar reports through RAG. It houses a collection of well-curated radiology reports encompassing detailed information on procedure names, findings, and impressions. For each report, the procedure name and findings are converted into vector representations. This transformation enables efficient retrieval and analysis of relevant reports, especially when the database is large.

\section{Experiments and Results}
\subsection{Dataset}
A total of 1,900 chest X-ray reports were collected from Massachusetts General Hospital between January 2018 and February 2018. Each report consists of three sections: procedure name, findings, and impression. From this dataset, 100 reports were randomly selected for model evaluation. The remaining 1,800 reports serve as the external database for RAG. The impression section in the raw dataset may contain information on report communication and management, which were detected and removed by an LLM (Llama-3.1-70b) \cite{dubey2024llama}. 
\subsection{Implementation Details}
Both the Radiologist and Reviewer agents are implemented by LLMs (Llama-3.1-70b) \cite{dubey2024llama} developed by Meta. Llama-3.1-70b is currently one of the leading open-source LLMs for text generation. The Report Retriever, a non-LLM agent, employs Faiss as the vector database, with vectorization performed by the GTE-base embedding model \cite{li2023towards}. The top 10 similar reports (i.e., \textit{k}=10) are retrieved from the external database based on the provided procedure name and findings. To prevent possible infinite review-revise loops and task drift, we limit the number of communication rounds to three. The system prompts for the Radiologist and Reviewer agents can be found in Appendix A.
\subsection{Memory Management}
Considering the limitations of the context window and the computational burden of LLMs, the memory capabilities of both the Radiologist and Reviewer agents are constrained. During the initial round of interaction, the radiologist retains the target procedure name and findings provided by the user, as well as the exemplar reports retrieved by the report retriever, to generate the initial impression. On the other hand, the Reviewer agent is only aware of the task provided by the user and the initial impression generated by the Radiologist agent for review. In subsequent rounds of interaction, the memory of both the Radiologist and the Reviewer agents is limited to their last round of communication and the task provided by the user. Exemplar reports retrieved earlier or previous rounds of conversations will not influence their ongoing interaction. 
\subsection{Performance of RadCouncil}
The performance of RadCouncil was evaluated using standard metrics for text generation tasks, including BLEU \cite{papineni2002bleu}, ROUGE-1, ROUGE-2, ROUGE-L \cite{lin2004rouge}, and BERTScore \cite{zhang2020bertscore}. These scores mainly focus on distance-based similarity (e.g., word matching) between the generated and ground truth impressions. The performance of RadCouncil is compared to a single-agent LLM (Llama-3.1-70b), which represents a radiologist operating without the support of exemplar reports or the reviewer. As listed in Table 2, RadCouncil outperformed the single-agent LLM across all these metrics. 

In addition to these metrics, RadCouncil was also evaluated using GPT-4o developed by OpenAI to evaluate, focusing on the semantic and clinical accuracy of the generated impression. The evaluation was based on the following qualitative criteria: \textit{inclusion of clinically significant findings}, \textit{consistency with original findings}, \textit{potential diagnosis}, \textit{stylistic concordance}, and \textit{conciseness and clarity}. Each criterion was scored on a scale from 1 to 10, where a higher score indicates better quality. The system prompts for evaluation can be found in Appendix B. The results of this evaluation are summarized in Table 3. The results show that both the single-agent and multi-agent systems performed equally well in identifying clinically significant findings (both 8.51 on a 1-10 scale)in the generated impression, demonstrating their ability to understand and process languages in radiology. In the task of inferring potential diagnoses, RadCouncil outperformed the single-agent LLM (8.24 vs. 8.15). RadCouncil also showed improvements in stylistic concordance (8.53 vs. 7.80) and conciseness/clarity (8.77 vs. 7.93) over single-agent LLM, possibly because of the support from retrieved exemplar reports. We also observed that the single-agent LLM achieved a better consistency score (8.77) than RadCouncil (8.63). We will analyze and discuss the inconsistency issue of RadCouncil in the next section. 

\begin{table}[h!]
\centering
\caption{Performance of RadCouncil compared with single LLMs by quantitative metrics}
\begin{tabular}{lccccc}
\hline
Model & BLEU & ROUGE\_1 & ROUGE\_2 & ROUGE\_L & BERTScore \\
\hline
Single LLM (Llama-3.1-70b) & 8.21 & 0.3490 & 0.1391 & 0.2745 & 0.6643 \\
RadCouncil & 24.22 & 0.5049 & 0.3088 & 0.4364 & 0.7434 \\
\hline
\end{tabular}
\end{table}

\begin{table}[h!]
\centering
\caption{The performance of multi-agent system compared with single agent using GPT-4 qualitative criteria.}
\begin{tabular}{lccccc}
\hline
Model & Findings & Consistency & Diagnosis & Style & Conciseness \\
\hline
Single LLM (Llama-3.1-70b) & 8.51 & 8.77 & 8.15 & 7.80 & 7.93 \\
RadCouncil & 8.51 & 8.63 & 8.24 & 8.53 & 8.77 \\
\hline
\end{tabular}
\end{table}

\subsection{Analysis of the RAG performance}
\begin{figure}[h]
\centering
\includegraphics[width=0.8\textwidth]{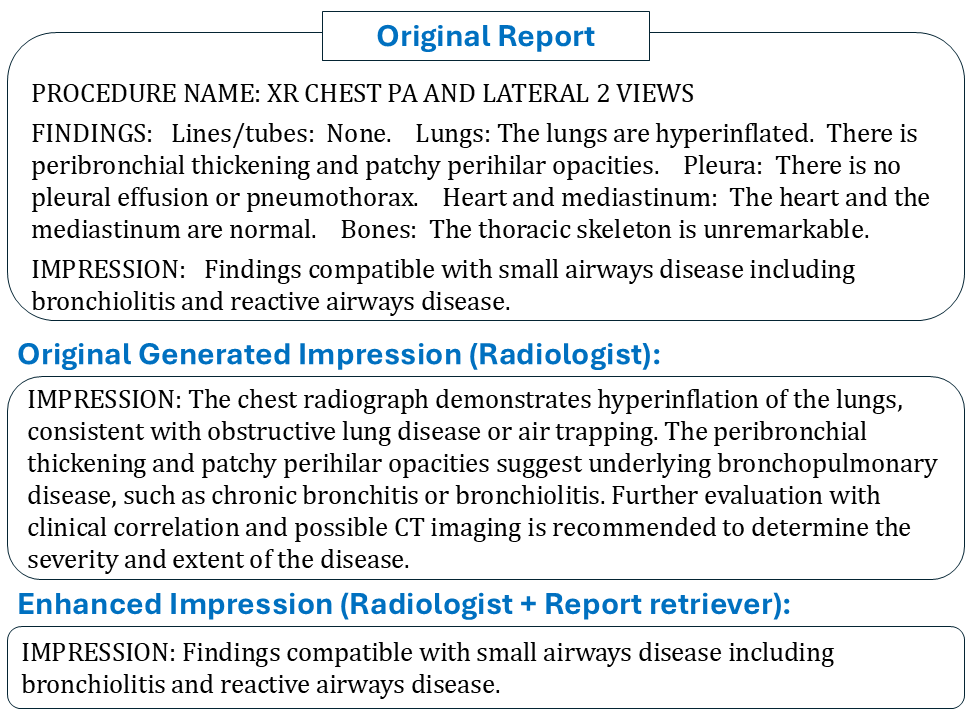}
\caption{Example comparison of original impression generated by radiologist agent only vs. impression enhanced by report retriever agent.}\label{fig2}
\end{figure}

Figure 2 shows an example of the original impression generated by the Radiologist agent alone and the enhanced impression with the support of the Retriever agent. This example demonstrates the benefit of RAG to make the generated impression more closely aligned with the ground truth. The enhanced impression shows improved stylistic concordance with typical radiological reporting practice. It is more concise and focused, eliminating extraneous details while retaining key findings. The accuracy of the potential diagnosis is also improved, as the final impression precisely identifies the condition as "small airways disease" and specifically mentions "bronchiolitis and reactive airways disease," perfectly aligning with the ground truth impression. With similar reports provided by the report retriever agent, the radiologist has the potential to generate more standardized and diagnostically accurate impressions.

On the other hand, providing additional information (in the form of exemplar reports) in the prompt to the Radiologist could result in inconsistent summaries in the generated impression. We manually reviewed the results of all 100 test cases, and analyzed five types of inconsistencies between the provided findings and generated impressions in various scenarios, as listed in Table 4. The five types of inconsistencies include Presence, Progression Status, Severity, Size, and Location of the findings. The "Presence" inconsistency refers to the cases where findings in the generated impression that are not mentioned in the input findings or the existence of the findings is contradictory to common sense. The "Progression Status" inconsistency indicates a mismatch between the temporal changes of lesion condition described in the findings and how they are represented in the impression (e.g., stating "worsening" in the impression when the findings indicate "stable" or "improved" conditions). The "Severity" inconsistency refers to cases where the degree or intensity of a finding is described differently. The "Size" inconsistency represents discrepancies in the dimensional descriptions of lesions. The "Location" inconsistency occurs when there are contradictions or mismatches in the anatomical location of findings. The analysis results show a notable increase in inconsistency of the impressions generated with the support from the Retriever agent (i.e., via RAG). In the impressions generated by the Radiologist agent alone, inconsistencies were relatively minimal, with only two inconsistencies (one on progression status and one on size) found in two cases. In contrast, impressions generated with RAG showed more inconsistencies across multiple categories. Among the 100 analyzed cases, ten of the impressions generated using RAG contained inconsistencies. There were seven instances of presence inconsistency, six instances of progression status inconsistency, one instance of severity inconsistency, and one instance of location inconsistency. Specifically, we found that RAG tends to generate unmentioned findings and add unmentioned progression status (especially “unchanged”) to the impression. This analysis demonstrates that while the use of RAG will enhance the overall quality of the generation, it could also introduce more inconsistencies, possibly due to the much longer context window for the exemplar reports. This suggests that further refinement and calibration for RAG would be necessary to reduce these inconsistencies and improve the reliability and trustworthiness.

\begin{table}[h!]
\centering
\caption{Summary of five types of inconsistencies in the generated impression by radiologist agent and the enhanced impression by report retriever.}
\begin{tabular}{lcccccc}
\hline
Model & Inconsistencies & Presence & Status & Severity & Size & Location \\
\hline
Single-agent (Radiologist) & 2 & 0 & 1 & 0 & 1 & 0 \\
Multi-agent & & & & & & \\
\quad + Retriever & 10 & 7 & 6 & 0 & 1 & 1 \\
\quad + Retriever \& Reviewer & 6 & 0 & 6 & 0 & 1 & 1 \\
\hline
\end{tabular}
\end{table}

\subsection{Effectiveness of the Reviewer agent}
To reduce inconsistencies in the generated impression, we implemented a Reviewer agent with prompt input consisting of only the original finding section and the generated impression. Our premise is that if the inconsistencies are caused by the long context window of the Radiologist agent, the Reviewer agent with a much simpler context can potentially identify the errors and make the corresponding revision suggestions. In this work, the Reviewer agent only focuses on the "Consistency" issue of the generated impression, which is the most critical problem of RadCouncil. As listed in the last row of Table 4, the Reviewer agent correctly identified and resolved inconsistencies in four cases, with most cases resolved in two rounds of interactions between the Radiologist and the Reviewer. The Reviewer correctly identified all the inconsistency errors regarding the Presence type, while it could not identify other types of inconsistencies. Specifically, regarding the "Status", the Reviewer agent failed to identify the wrongfully-generated “unchanged” conclusions in the impression. It explained that the absence of a changed status implies the condition is unchanged, which is incorrect according to radiological reporting standards, as the "unchanged" status must be explicitly supported by prior comparison studies mentioned in the findings. This limitation suggests that the Reviewer agent may benefit from additional training specifically on temporal relationships and status changes in radiological findings. 

\section{Conclusion and Discussion}
This study presents RadCouncil, a multi-agent framework for generating radiology report impressions based on the input finding section and procedure name. Experiment results demonstrate significant improvements over single-agent approaches. Through quantitative metrics (BLEU, ROUGE, BERTScore) and qualitative evaluation by GPT-4, RadCouncil showed enhanced performance in diagnostic accuracy, stylistic concordance, and clarity of expression, aligning with recent findings by Sun et al. \cite{sun2023evaluating} on the potential of LLMs in radiology workflows. As healthcare systems continue to face increasing demands and workload pressures \cite{maskell2022does}, AI-assisted documentation tools such as the RadCouncil developed in this work could play a vital role in maintaining high-quality patient care while supporting healthcare provider efficiency.

The three-agent architecture of RadCouncil, combining a Report Retriever for similar case identification, a Radiologist for impression generation, and a Reviewer for consistency checking, proved effective in producing more standardized and accurate impressions. This success supports the emerging consensus that multi-agent systems can effectively handle complex medical tasks through specialized role distribution, as shown in similar healthcare applications \cite{yue2024ct,ke2024enhancing,wei2024medco}.

The demonstrated effectiveness of RadCouncil, despite its limitations, suggests potential applications beyond radiology to other healthcare domains where complex task coordination and specialized expertise are crucial for accurate documentation and decision-making. We envision that multi-agent systems will be increasingly utilized in healthcare AI development, especially those involving sophisticated reasoning and domain expertise \cite{tang2023medagents, smit2023we}.

We found that the retrieval-augmented generation (RAG) approach particularly improved the stylistic alignment with radiological reporting conventions and diagnostic precision, extending the findings of Ma et al. \cite{ma2024iterative}. However, our analysis revealed important challenges in using RAG, as it could introduce additional inconsistencies. This aligns with concerns raised by Omiye et al. \cite{omiye2024large} regarding LLM hallucinations in medical contexts. To mitigate this problem, we implemented the Reviewer agent, which showed promising performance in identifying and correcting such inconsistencies.
Findings from this study suggest several promising avenues for advancing both radiology applications and broader healthcare applications of multi-agent systems: 1) Further improvements of the Radiologist and Reviewer agent's capabilities, particularly in temporal reasoning and status change detection, are much needed. This could build upon the analysis of longitudinal radiology reports by these agents similar to the clinical practice. 2) To address the context window limitations in our current RAG implementation that can lead to information overflow and inconsistencies, future work is needed to develop more advanced memory management techniques such as hierarchical retrieval strategies and dynamic context prioritization \cite{zhang2024hierarchical, su2024dragin}.

\bibliography{LLM_refs}
\bibliographystyle{unsrt}
\newpage
\appendix

\section*{Appendix A.\\System prompts for the Radiologist and Report Retriever agents}
\addcontentsline{toc}{section}{AppendixA} 

\textbf{Radiologist:}\\
\begin{quote}
You are a thoracic radiologist. Based on the given 'FINDINGS' section of the chest image report, your task is to derive the 'IMPRESSION' section which contains summarization of the clinically significant findings and possible diagnostic interpretations. If reviewers' feedback is provided, consider it as an additional context to revise your impression. Ensure that your derived 'IMPRESSION' is clear, concise, accurate. Start your response directly with 'IMPRESSION:' followed by your impression, without any preamble or additional comments or notes.
\end{quote}

\textbf{Report retriever:}\\
\begin{quote}
You are a reviewer for the 'IMPRESSION' section in chest image reports. Your task is to verify whether each finding mentioned in the 'IMPRESSION' section is consistent with the 'FINDINGS' section. Specifically, focus on the details regarding size, location, severity, and progression status of the findings.
If you identify any inconsistent descriptions in the 'IMPRESSION', end your response with ‘REVISION: YES’.
If all findings mentioned in the 'IMPRESSION' are consistent, end your response with ‘REVISION: NO’.
\end{quote}

\section*{Appendix B.\\System prompt used for evaluation by GPT-4o}
\addcontentsline{toc}{section}{AppendixA} 
\begin{quote}
You are an expert thoracic radiologist with extensive experience in your field. Your task is to evaluate and score the quality of generated impressions for chest image reports produced by large language models (LLMs). Using the findings from the original report as context, compare the quality of each generated impression with the original impression. You will be provided with an original report, including both the findings and the original impression, as well as generated impressions from two models: a single-agent model and a multi-agent model.
Your evaluation should be based on the following five factors. Assign a score from 0 to 10 for each factor, where a higher score indicates better quality.

\begin{enumerate}
    \item Clinically Significant Findings: Evaluate whether any clinically significant findings present in the original impression are missing from the generated impression.
    \item Consistency with Original Findings: Assess how well the generated impression aligns with the findings detailed in the original report.
    \item Potential Diagnosis: Determine whether the generated impression appropriately suggests or addresses potential diagnoses based on the findings and whether these align with the original impression.
    \item Stylistic Concordance: Compare the style of the generated impression with that of the original impression.
    \item Conciseness and Clarity: Judge the generated impression on its clarity and conciseness relative to the original impression.
\end{enumerate}

Be as objective as possible in your assessment. Output your evaluation in JSON format as follows, providing concise explanations for each score: \\
\{ \\
   \hspace*{0.5cm} "Single\_agent\_model": \{ \\
   \hspace*{1cm} "Significant\_findings": \{ "Score": <score>, "Reason": "<explanation>" \}, \\
   \hspace*{1cm} "Consistency": \{ "Score": <score>, "Reason": "<explanation>" \}, \\
   \hspace*{1cm} "Diagnosis": \{ "Score": <score>, "Reason": "<explanation>" \}, \\
   \hspace*{1cm} "Style": \{ "Score": <score>, "Reason": "<explanation>" \}, \\
   \hspace*{1cm} "Conciseness\_and\_clarity": \{ "Score": <score>, "Reason": "<explanation>" \} \\
  \hspace*{0.5cm}\}, \\
  \hspace*{0.5cm}"Multi\_agent\_model": \{ \\
    \hspace*{1cm} "Significant\_findings": \{ "Score": <score>, "Reason": "<explanation>" \}, \\
    \hspace*{1cm} "Consistency": \{ "Score": <score>, "Reason": "<explanation>" \}, \\
    \hspace*{1cm} "Diagnosis": \{ "Score": <score>, "Reason": "<explanation>" \}, \\
    \hspace*{1cm} "Style": \{ "Score": <score>, "Reason": "<explanation>" \}, \\
    \hspace*{1cm} "Conciseness\_and\_clarity": \{ "Score": <score>, "Reason": "<explanation>" \} \\
  \hspace*{0.5cm}\} \\
\} \\
\end{quote}
\end{document}